\documentclass{article}
\usepackage{spconf,amsmath,graphicx}

\usepackage{times}
\usepackage{latexsym}
\usepackage[T1]{fontenc}
\usepackage{microtype}
\usepackage{inconsolata}

\usepackage{epigraph}
\usepackage{graphicx}
\usepackage{amsmath}
\usepackage{amssymb}
\usepackage{mathtools}
\usepackage{booktabs}
\usepackage{algorithm}
\usepackage{algorithmic}
\usepackage{wrapfig}
\usepackage{glossaries}
\usepackage{colortbl}
\usepackage{bbding}
\usepackage{tikz}
\usepackage{comment}
\usepackage{color}
\usepackage{xspace}
\usepackage{booktabs}
\usepackage{nicefrac}
\usepackage{bbm}
\usepackage{bm}
\usepackage{tabularx,verbatim}
\usepackage{multirow}
\usepackage{subcaption}
\usepackage{array}
\usepackage[accsupp]{axessibility}  
\usepackage{pifont}
\usepackage{tikz}
\usepackage{newfloat}
\usepackage{listings}
\definecolor{remark}{rgb}{1,.5,0} 
\definecolor{cyan}{rgb}{0.831,0.901,0.945}

\floatstyle{ruled}
\newfloat{listing}{tb}{lst}{}
\floatname{listing}{Listing}

\usepackage{amsmath,amsfonts,bm}

\def\eqref#1{equation~\ref{#1}}

\def\1{\bm{1}}

\DeclareMathAlphabet{\mathsfit}{\encodingdefault}{\sfdefault}{m}{sl}
\SetMathAlphabet{\mathsfit}{bold}{\encodingdefault}{\sfdefault}{bx}{n}

\DeclareMathOperator*{\argmax}{arg\,max}

\definecolor{me_node}{RGB}{16,119,51}
\definecolor{sup_node}{RGB}{0, 90, 181}
\definecolor{sub_node}{RGB}{212, 17, 89}

\definecolor{correct_dts}{RGB}{45, 176, 117}
\definecolor{wrong_dts}{RGB}{255, 138, 216}

\definecolor{upperboundcolor}{gray}{.5}
\definecolor{vanillacolor}{RGB}{238, 232, 170}

\definecolor{higher}{RGB}{16,119,51}
\definecolor{lower}{RGB}{220,50,32}

\definecolor{ours_meth}{RGB}{16,119,51}
\definecolor{bl_meth}{RGB}{115,144,167}

\definecolor{highlightGreen}{rgb}{64, 252, 167}
\definecolor{promptGreen}{rgb}{0.251, 0.992, 0.655}

\definecolor{firstBest}{rgb}{0.93, 1, 0.93}
\definecolor{secondBest}{rgb}{0.93, 0.93, 1}

\definecolor{baselinecolor}{gray}{.9}

\newcommand{\fbest}[1]{\cellcolor{firstBest}{#1}}
\newcommand{\sbest}[1]{\cellcolor{secondBest}{#1}}

\newcolumntype{x}[1]{>{\centering\arraybackslash}p{#1pt}}
\newcolumntype{y}[1]{>{\raggedright\arraybackslash}p{#1pt}}
\newcolumntype{z}[1]{>{\raggedleft\arraybackslash}p{#1pt}}

\newlength\savewidth

\newcommand{\vocada}{VocAda\xspace}
\newcommand{\vocadaFull}{Vocabulary Adapter\xspace}

\newcommand{\ovod}{OvOD\xspace}
\newcommand{\shine}{SHiNe\xspace}

\newcommand{\vlm}{VLM\xspace}

\newcommand{\llm}{LLM\xspace}

\newcommand{\enctxt}{\mathcal{E}_{\text{txt}}}

\newcommand{\bI}{\mathbf{I}}
\newcommand{\bb}{\mathbf{b}}
\newcommand{\bB}{\mathbf{B}}

\newcommand{\bz}{\mathbf{z}}

\newcommand{\classes}{\mathcal{C}}

\newcommand{\captions}{\mathcal{S}}
\newcommand{\nouns}{\mathcal{P}}

\definecolor{DarkCoral}{rgb}{0.8, 0.36, 0.27}

\usepackage{url}

\usepackage{hyperref}

\usepackage[capitalize]{cleveref}
\crefname{section}{Sec.}{Secs.}
\Crefname{section}{Section}{Sections}
\Crefname{table}{Table}{Tables}
\crefname{table}{Tab.}{Tabs.}

\newcommand{\lblsec}[1]{\label{sec:#1}}
\newcommand{\lblfig}[1]{\label{fig:#1}}
\newcommand{\lbltab}[1]{\label{tbl:#1}}

\newcommand{\lbleq}[1]{\label{eq:#1}}

\newcommand{\refsec}[1]{Sec.~\ref{sec:#1}}
\newcommand{\reffig}[1]{Fig.~\ref{fig:#1}}
\newcommand{\reftab}[1]{Tab.~\ref{tbl:#1}}

\newcommand{\refapp}[1]{Supp.~\ref{sec:#1}}

\newcommand{\myparagraph}[1]{\vspace{0.05cm}\noindent\textbf{#1}}



\newcommand{\lesspace}{{\vspace{-0.4cm}}}

\definecolor{baseline_fig1}{RGB}{194, 46,  90}
\definecolor{ours_fig1}{RGB}{37, 89,  175}

\definecolor{butter1}{RGB}{252, 233,  79}
\definecolor{butter2}{RGB}{237, 212,   0}
\definecolor{butter3}{RGB}{196, 160,   0}
\colorlet{LightButter}{butter1}
\colorlet{Butter}{butter2}
\colorlet{DarkButter}{butter3}

\definecolor{orange1}{RGB}{252, 175,  62}
\definecolor{orange2}{RGB}{245, 121,   0}
\definecolor{orange3}{RGB}{206,  92,   0}
\colorlet{LightOrange}{orange1}
\colorlet{Orange}{orange2}
\colorlet{DarkOrange}{orange3}

\definecolor{chocolate1}{RGB}{233, 185, 110}
\definecolor{chocolate2}{RGB}{193, 125,  17}
\definecolor{chocolate3}{RGB}{143,  89,   2}
\colorlet{LightChocolate}{chocolate1}
\colorlet{Chocolate}{chocolate2}
\colorlet{DarkChocolate}{chocolate3}

\definecolor{chameleon1}{RGB}{138, 226,  52}
\definecolor{chameleon2}{RGB}{115, 210,  22}
\definecolor{chameleon3}{RGB}{ 78, 154,   6}
\colorlet{LightChameleon}{chameleon1}
\colorlet{Chameleon}{chameleon2}
\colorlet{DarkChameleon}{chameleon3}

\definecolor{skyblue1}{RGB}{114, 159, 207}
\definecolor{skyblue2}{RGB}{ 52, 101, 164}
\definecolor{skyblue3}{RGB}{ 32,  74, 135}
\colorlet{LightSkyBlue}{skyblue1}
\colorlet{SkyBlue}{skyblue2}
\colorlet{DarkSkyBlue}{skyblue3}

\definecolor{plum1}{RGB}{173, 127, 168}
\definecolor{plum2}{RGB}{117,  80, 123}
\definecolor{plum3}{RGB}{ 92,  53, 102}
\colorlet{LightPlum}{plum1}
\colorlet{Plum}{plum2}
\colorlet{DarkPlum}{plum3}

\definecolor{scarletred1}{RGB}{239,  41,  41}
\definecolor{scarletred2}{RGB}{204,   0,   0}
\definecolor{scarletred3}{RGB}{164,   0,   0}
\colorlet{LightScarletRed}{scarletred1}
\colorlet{ScarletRed}{scarletred2}
\colorlet{DarkScarletRed}{scarletred3}

\definecolor{aluminium1}{RGB}{238, 238, 236}
\definecolor{aluminium2}{RGB}{211, 215, 207}
\definecolor{aluminium3}{RGB}{186, 189, 182}
\definecolor{aluminium4}{RGB}{136, 138, 133}
\definecolor{aluminium5}{RGB}{ 85,  87,  83}

\definecolor{indigo}{RGB}{114,  33, 188}
\definecolor{maroon}{RGB}{103,   7,  72}
\definecolor{turquoise}{RGB}{ 64, 224, 208}
\definecolor{green4}{RGB}{  0, 139,   0}

\crefname{section}{\S}{\S\S}
\crefname{subsection}{\S}{\S\S}
\crefformat{table}{Table~#2#1#3}
\crefformat{figure}{Fig.~#2#1#3}
\crefformat{equation}{Eq.~#2#1#3}
\newcommand{\rowNumber}[1]{}
\definecolor{highlightRowColor}{rgb}{0.95, 0.95, 1}
\definecolor{baselineRowColor}{rgb}{0.95, 0.95, 0.95}

\definecolor{correctpred}{RGB}{0,153,77}
\definecolor{makesensepred}{RGB}{209, 70, 153}
\definecolor{wrongpred}{RGB}{250, 0, 0}
\definecolor{superpred}{RGB}{0, 127, 255}

\def\eg{\emph{e.g.\,}}

\def\maketitlesupplementary
   {
   \newpage
       \twocolumn[
        \centering
        \Large
        \textbf{Test-time Vocabulary Adaptation for Language-driven Object Detection}\\
        \vspace{0.5em}Supplementary Material \\
        \vspace{1.0em}
       ] 
   }

\title{Test-time Vocabulary Adaptation for Language-driven Object Detection}
\name{
\begin{tabular}{ccc}
Mingxuan Liu$^{1,*}$\thanks{$^{*}$Work done at NAVER LABS Europe.}
& Tyler L. Hayes$^{2}$ & Massimiliano Mancini$^{1}$ \\
Elisa Ricci$^{1,3}$ & Riccardo Volpi$^{4,*}$ & Gabriela Csurka$^{2}$
\end{tabular}
}
\address{
$^{1}$University of Trento \quad $^{2}$NAVER LABS Europe \quad $^{3}$Fondazione Bruno Kessler \quad $^{4}$Arsenale Bioyards
}

\begin{document}
%
\maketitle

\begin{abstract}
Open-vocabulary object detection models 
{allow}
users to freely specify a class vocabulary in natural language at test time, guiding the detection of desired objects. However, vocabularies can be overly broad or even mis-specified, hampering the overall performance of the detector. In this work, we propose a \textit{plug-and-play} Vocabulary Adapter (VocAda) to refine the user-defined vocabulary, automatically tailoring it to categories that are relevant for a given image. VocAda {does not require any training,} {it} operates at inference {time} in three steps: \textit{i)} it uses an image captionner to describe visible objects, \textit{ii)} {it} parses nouns from those captions, and \textit{iii)} {it} selects relevant classes from the user-defined vocabulary, discarding irrelevant ones.
Experiments on COCO and Objects365 with three state-of-the-art detectors show that VocAda consistently 
{improves}
performance, 
proving its {versatility}.
The code is \href{https://github.com/OatmealLiu/VocAda}{open source}.
\end{abstract}

\section{Introduction}
\lblsec{introduction}

{The goal of object detection is answering the question \emph{“What objects are present, and where?”}, by locating and classifying objects in images.}
Such detection is vital for applications like autonomous driving and embodied AI~\cite{ramrakhya2024seeing}. Traditional detectors~\cite{he2017mask,redmon2018yolov3} only recognize classes seen in training, requiring 
{finetuning}
for new classes. To overcome this limitation, \textit{open-vocabulary} object detection (\ovod) uses contrastive vision-language models~\cite{radford2021learning} that align visual and textual representations in a joint space. This alignment lets \ovod detectors handle user-defined vocabularies without retraining, enabling flexible adaptation to specific interests or applications.

While the \ovod paradigm supports zero-shot detection of arbitrary concepts, overly broad user-defined vocabularies can introduce noise and hurt performance. For instance, a state-of-the-art \ovod detector like Detic~\cite{zhou2022detecting} may mistake a ``Curling'' stone for a ``Teapot'', as shown in \reffig{main_qualitative}, due to their visual similarity ({\includegraphics[height=1em]{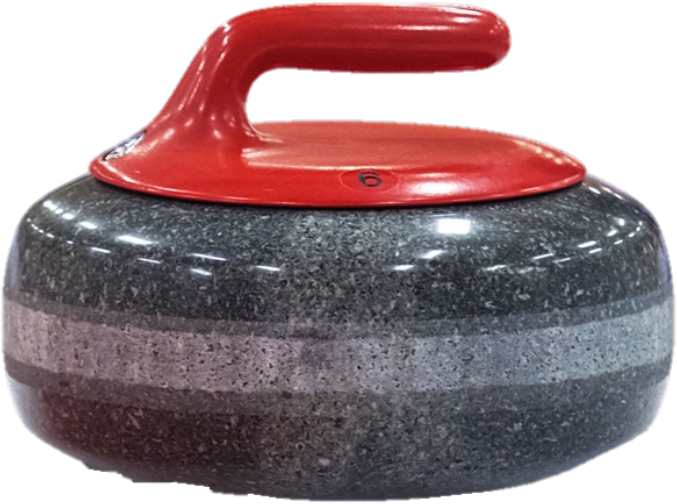}} \textit{vs.} {\includegraphics[height=1em]{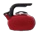}}). This happens because most \ovod methods rely on region-class similarity without leveraging the full scene context. A model that interprets the entire scene is far less likely to misclassify sports equipment as a ``Teapot''.

\begin{figure}[t]
  \begin{center}
  \includegraphics[width=0.9\columnwidth]{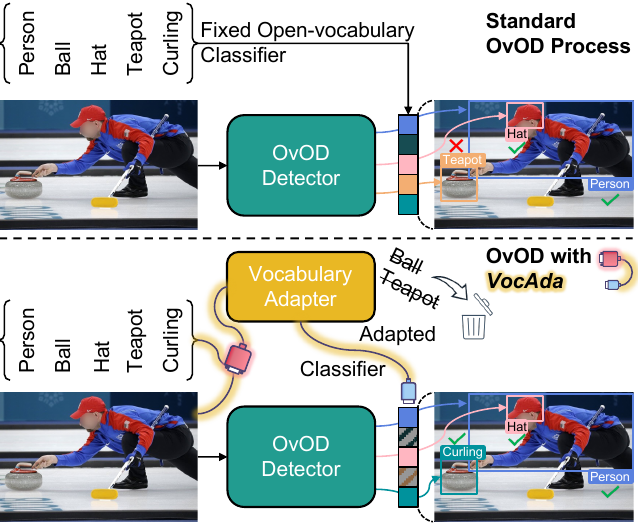}
  \end{center}
    \vspace{-6.0mm}

  \caption{
  \textbf{Top}: Standard \ovod methods continuously operate on a fixed, noisy vocabulary defined by the user.
  \textbf{Bottom}: We propose a \textit{plug-and-play} Vocabulary Adapter (VocAda) to adapt this vocabulary to the current image based on the semantic interpretation of visual content during inference, improving the detector's performance by discarding non-relevant classes.
  }
  \vspace{-6.0mm}
  \lblfig{teaser}
\end{figure}

To address this problem, we propose \vocadaFull (\vocada), a training-free module that refines user-defined vocabularies using semantic context at inference time, boosting the performance of off-the-shelf \ovod detectors. As shown in \reffig{teaser}, \vocada filters out irrelevant classes (distractors) so that the detector’s classifier focuses only on those 
{actually}
present. The main challenge lies in pinpointing classes both relevant to the user and visible in the image. We address this by leveraging LLaVA-Next~\cite{liu2024llavanext}, a vision-language model (VLM) that generates a list of objects when prompted, and then parsing noun phrases (see \reffig{teaser_secondary}). Because the extracted nouns often differ from user-defined class names (\eg, ``Riders'' \textit{vs.}\ ``Person''), \vocada provides two class-selection strategies: one using text similarity, and the other relying on a large language model (\llm) to link the parsed nouns to vocabulary classes.

\vocada 
can be seamlessly integrated
into any \ovod detector, 
{and runs in parallel to it,}
minimizing the computational overhead introduced by large VLMs. In our experiments, we apply it to three \ovod detectors: Detic~\cite{zhou2022detecting}, VLDet~\cite{lin2022learning}, and CoDet~\cite{ma2024codet}, evaluating on COCO~\cite{lin2014microsoft} and Objects365~\cite{shao2019objects365}. \vocada consistently improves performance across all detectors and benchmarks, highlighting new opportunities in context-driven vocabulary adaptation.

\begin{figure}[t]
  \begin{center}
  \includegraphics[width=0.85\columnwidth]{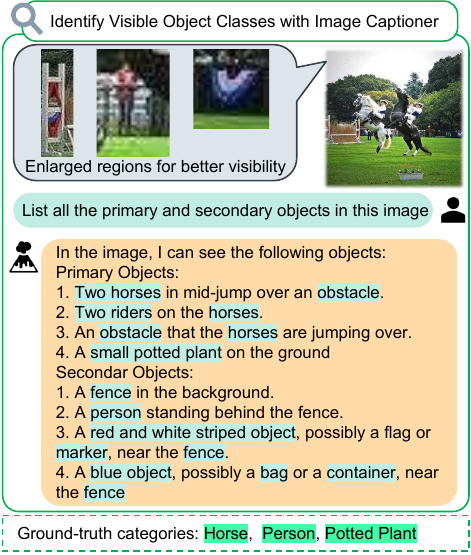}
  \end{center}
  \lesspace
    \caption{
      The image caption obtained with LLaVA-Next-7B demonstrates impressive comprehensiveness in identifying visible object classes from the image. The ground-truth classes from the user-defined vocabulary that appear in the image are listed at the bottom. 
    }

  \lesspace
  \lblfig{teaser_secondary}
\end{figure}

To summarize, our key contributions are:
\textit{i)} highlighting the need for an image-dependent vocabulary in \ovod, and showing that an optimal vocabulary (oracle) yields significant improvements;
\textit{ii)} proposing \vocada, a model-agnostic and training-free method for vocabulary adaptation in \ovod;
\textit{iii)} conducting extensive experimental validation, showcasing the versatility of our proposed solution across various detectors and benchmarks. The code will be released upon publication.
\section{Related Work}
\lblsec{relatedwork}

Open-vocabulary object detection (\ovod)~\cite{zhu2023survey} aims to map predicted region features to a frozen vision-language embedding space, typically from contrastive models like CLIP~\cite{radford2021learning}. \ovod detectors usually train on box-labeled data~\cite{lin2014microsoft,gupta2019lvis} with limited categories due to high annotation costs, and supplement these with 
datasets
{annotated at image level}~\cite{deng2009imagenet}, which cover more classes. Major studies have focused on improving alignment training via pseudo-labeling~\cite{zhou2022detecting}, transfer learning~\cite{zhong2022regionclip}, or enhanced weak supervision~\cite{ma2024codet}. In contrast, we improve off-the-shelf \ovod detectors without fine-tuning, 
{updating}
only their vocabularies. Our work relates to SHiNe~\cite{liu2024shine}, which augments vocabularies via prompt engineering and a semantic hierarchy, but produces a single improved vocabulary for all images. Instead, \vocada adapts the vocabulary per image at test time, complementing prompt-engineering approaches that can further augment \vocada's refined vocabulary.

\section{Method}
\lblsec{mthd_method}
\begin{figure*}[t]
  \centering
  \includegraphics[width=0.94\linewidth]{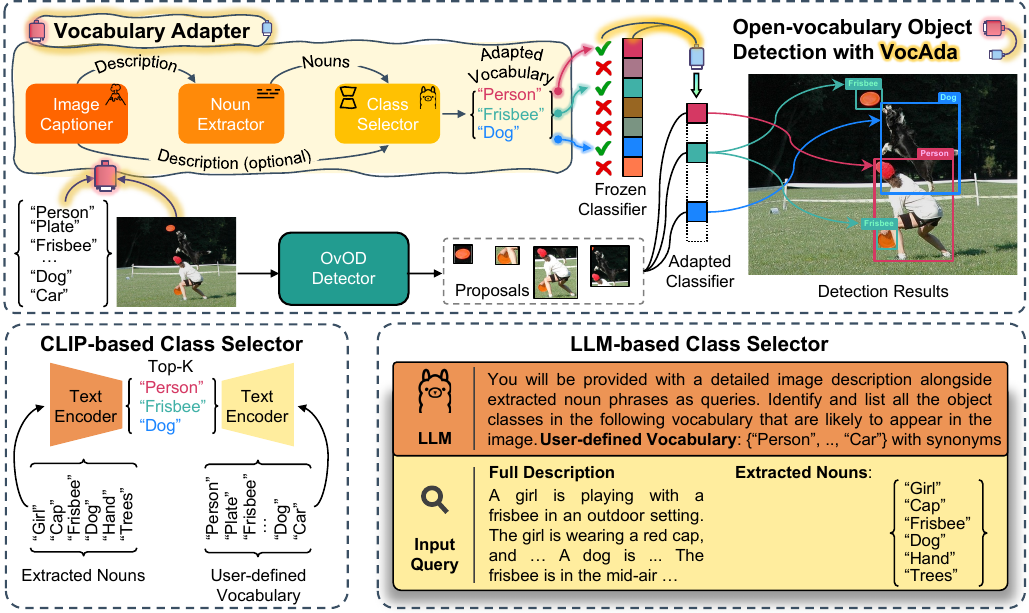}
  \vspace{-3.0mm}
  \caption{%
  \textbf{Pipeline of open-vocabulary object detection with \vocada.}
  \textbf{Top:} Given an image and a user-defined vocabulary, \vocada uses an Image Captioner to describe visible objects, then extracts nouns representing category names via a Noun Extractor. Next, the Class Selector matches these nouns to the user vocabulary, yielding an adapted vocabulary for final detection.
  \textbf{Bottom:} Two Class Selector variants:
  (\textit{left}) top-$k$ CLIP text similarity; 
  (\textit{right}) an \llm{} that proposes the adapted vocabulary.
  }%
  \lblfig{system}
  \vspace{-6.0mm}
\end{figure*}

In this work, we aim to improve off-the-shelf \ovod detectors via \textit{vocabulary adaptation}. First, we briefly introduce modern \ovod models. Then, we present \vocada, detailing its key components and their integration in a detection pipeline.

\myparagraph{Background.}
Modern \ovod detectors~\cite{zhu2023survey} follow a two-stage process: \textit{Stage 1:} A region proposal network takes an image $\bI$ and outputs region proposals $\{(\bb_n, \bz_n)\}_{n=1}^{N_b}$, where each $\bb_n \in \mathbb{R}^{4}$ is a predicted bounding box and $\bz_n \in \mathbb{R}^D$ is its $D$-dimensional embedding in a pre-trained vision-language space (\eg, CLIP~\cite{radford2021learning}); \textit{Stage 2:} Each proposal can optionally refine its box coordinate $\bb_n \leftarrow \bB\bz_n + \bb_n$ via a learned projector $\bB \in \mathbb{R}^{4 \times D}$. A set of user-defined class names $\classes=\{c_j\}_{j=1}^{N_c}$ (\textit{vocabulary}) is then used to classify each proposal:
\begin{equation}
\vspace{-1.0mm}
y_n = \argmax_{c \in \classes} \bigl\{ \enctxt(\text{T}(c)) \cdot \bz_n \bigr\},
\lbleq{ovod}
\vspace{-1.0mm}
\end{equation}
where $\enctxt$ is a frozen CLIP text encoder, $\text{T}(c)$ is a prompt (e.g., ``a \{Class Name\}'') for the class name, and $\bz_n$ is the proposal’s visual embedding. During training, $\enctxt$ remains frozen while the detector learns to align $\bz_n$ with its corresponding text representation $\enctxt(\text{T}(c_n))$, where $c_n$ is the ground-truth class name. Once trained \ovod models let users freely specify $\classes$ at test time--including novel classes unseen during training--and use \cref{eq:ovod} to assign labels in an open-vocabulary setting.

\myparagraph{OvOD with \vocada.} 
{Prior {related} works 
{focus on better aligning}
the regional feature $\bz_n$ with the vision-language embedding space~\cite{zhu2023survey} or refin{ing} the prompt $\text{T}(\cdot)$ using hierarchical semantics~\cite{liu2024shine}. In contrast, }we \textit{actively adapt the vocabulary} $\classes$ to each image by removing irrelevant classes (distractors), reducing ambiguity and 
{enhancing}
recognition quality.
Specifically, as illustrated in \reffig{system}, given an image $\bI$ and a user-defined vocabulary $\classes$, we keep the \ovod module frozen and integrate our \vocada to adapt $\classes$ into a subset $\widetilde{\classes}_{I} \subseteq \classes$. This vocabulary adaptation process identifies the relevant categories present in the image while discarding as many irrelevant ones as possible. Consequently, it guides \cref{eq:ovod} to classify proposals exclusively within $\widetilde{\classes}_{I}$ as:
\begin{equation}
\vspace{-1.0mm}
y_n = \argmax_{c \in \widetilde{\classes}_{I}}
\bigl\{ \enctxt(\text{T}(c)) \cdot \bz_n \bigr\}.
\lbleq{vocada_assignment}
\vspace{-1.0mm}
\end{equation}

\subsection{Test-time Vocabulary Adaptation}
As shown in \reffig{system}, \vocada consists of three modules:
\textit{(i)} an \textit{Image Captioner}, a VLM that generates a natural language description of visible object categories in the image $\bI$;
\textit{(ii)} a \textit{Noun Extractor}, which parses this description to extract noun phrases;
\textit{(iii)} a \textit{Class Selector}, which takes the extracted nouns, the user-defined vocabulary $\classes$, and optionally the description to produce a subset $\widetilde{\classes}_{I} \subset \classes$ that are relevant to the current image and of interest to the users ($\widetilde{\classes}_{I} \subset \classes$).

\myparagraph{Image Captioner (IC).}
The goal of the \textbf{IC} is to generate an accurate and comprehensive description $\captions_{I}$ of the objects that are visible in a given image.
We prompt LLaVA-Next-7B~\cite{liu2024llavanext} to generate a detailed description $\captions_{I}$ of visible objects in an image. To ensure comprehensive coverage of all objects in the image, we prompt the \textbf{IC} to list \textit{primary} (large or foreground) and \textit{secondary} (small or background) objects, respectively. This prompt design effectively guides the model to interpret and capture all objects in the scene. The full prompt is in \refapp{prompt_ic}. Other VLMs~\cite{ye2023mplug2} could also replace LLaVA as IC.

\myparagraph{Noun Extractor (NE).}
The \textbf{NE} module isolates category information we need to adapt the vocabulary by extracting noun phrases $\nouns_{I} = \{p_{m}\}_{m=1}^{N_p}$ from the \textbf{IC}'s description $\captions_{I}$.
We use spaCy~\cite{Honnibal_spaCy_Industrial} to tokenize the text, assign part-of-speech tags, and perform dependency parsing, producing \textit{n}-gram noun phrases (\eg ``plastic containers'' instead of just ``containers'') to retain context-relevant adjectives for the subsequent selection step.

\myparagraph{Class Selector (CS).}
Given the extracted nouns $\nouns_{I}$ and the user-defined vocabulary $\classes$, the CS identifies a subset $\widetilde{\classes}_{I} \subset \classes$ relevant to the current image, discarding non-appearing or distracting classes. We propose two \textbf{CS} variants:

\noindent\textit{i) CLIP-based CS:}
We embed $\nouns_{I}$ and $\classes$ with CLIP ViT-L/14 and, for each noun phrase $p_m$, we select the top-$k$ most similar classes $\widetilde{\classes}_{p_m}=\{\tilde{c}_1, \dots, \tilde{c}_k\}$ from the vocabulary $\classes$ by text similarity.
The union of these sets forms the adapted vocabulary as $\widetilde{\classes}_{I} = \bigcup_{m=1}^{N_{p}} \widetilde{\classes}_{p_m}$.
We set $k=1$ and analyze 
{other choices}
in \refsec{expt_analysis}.
\noindent\textit{(ii) \llm-based} \textbf{CS:} 
Although the CLIP-based \textbf{CS} can effectively match extracted noun phrases to corresponding classes, its word-to-word mechanism can fail when nouns or class names are ambiguous. For example, if $p_m = \text{``Bat''}$ is extracted from an image of a baseball player using a bat, and both ``Bat'' (the animal) and ``Baseball Bat'' are in $\classes$, the former may rank higher but be \textit{incorrect}. To address this, we propose an alternative \textbf{CS} that leverages the \llm{}’s explicit context-aware reasoning. Specifically, we embed a task instruction and the vocabulary $\classes$, enriched with synonyms, into the \llm{}’s system prompt. These synonyms, queried in advance from an \llm{}, help recognize variant phrasings (\eg, ``TV'' \textit{vs.} ``Television'') and thus improve selection quality (see \refapp{additional_expt}). During inference, the \llm{} processes the entire description $\captions_{I}$ and the extracted nouns $\nouns_{I}$, then automatically proposes the subset $\widetilde{\classes}_{I}$ from the user-defined $\classes$. We use Llama3-8B~\cite{meta2024llama3} as our \llm{}
{(}see prompt details in \refapp{prompt_llm}{)}.
\begin{table*}[t]
    \centering
    \small
    \setlength{\tabcolsep}{4.8pt} 
        \caption{
    \textbf{Comparison on OVE-COCO.} 
    We apply \vocada to three \ovod detectors (Detic, VLDet, and CoDet) and compare its variants with the {\colorbox{vanillacolor}{baseline}} and the {\color{upperboundcolor}{\vocada-Oracle}} method. All use a ResNet-50 backbone. We report AP$_{50}$ and the average gap ($\overline{\Delta}$) versus the baseline, averaged across detectors. 
    {\colorbox{firstBest}{Best}} and {\colorbox{secondBest}{second-best}} results are highlighted.
    }
    \vspace{-2.0mm} 
    \begin{tabular}{lcccccccccccc}
        \toprule
        \multicolumn{1}{c}{\multirow{2}{*}{Method}}
        & \multicolumn{3}{c}{Detic} 
        & \multicolumn{3}{c}{VLDet}
        & \multicolumn{3}{c}{CoDet}
        & \multicolumn{3}{c}{$\overline{\Delta}$} \\

        \cmidrule(lr){2-4}
        \cmidrule(lr){5-7}
        \cmidrule(lr){8-10}
        \cmidrule(lr){11-13}

        & $\text{AP}_{50}^{novel}$ & $\text{AP}_{50}^{base}$ & $\text{AP}_{50}^{all}$
        & $\text{AP}_{50}^{novel}$ & $\text{AP}_{50}^{base}$ & $\text{AP}_{50}^{all}$
        & $\text{AP}_{50}^{novel}$ & $\text{AP}_{50}^{base}$ & $\text{AP}_{50}^{all}$
        & $\text{AP}_{50}^{novel}$ & $\text{AP}_{50}^{base}$ & $\text{AP}_{50}^{all}$ \\


        \cmidrule{1-1}
        \cmidrule(lr){2-4}
        \cmidrule(lr){5-7}
        \cmidrule(lr){8-10}
        \cmidrule(lr){11-13}
        
        \rowcolor{vanillacolor}{Baseline}
        & {27.8} & {51.1} & {45.0} 
        & {32.0} & {50.6} & {45.7} 
        & {30.5} & {52.5} & {46.7} 
        & {-} & {-} & {-} \\

        \cmidrule{1-1}
        \cmidrule(lr){2-4}
        \cmidrule(lr){5-7}
        \cmidrule(lr){8-10}
        \cmidrule(lr){11-13}

        w. RAM++
        & {29.2} & {47.4} & {42.7}
        & {33.7} & {47.0} & {43.5} 
        & {32.1} & {48.8} & {44.5} 
        & {\color{higher}+1.6} & {\color{lower}-3.6} & {\color{lower}-2.3} \\
        
        \cmidrule{1-1}
        \cmidrule(lr){2-4}
        \cmidrule(lr){5-7}
        \cmidrule(lr){8-10}
        \cmidrule(lr){11-13}
        
        w. \vocada-SBert
        & {29.6} & {50.0} & {44.7} 
        & {34.4} & {49.7} & {45.7} 
        & {33.2} & {51.8} & {46.9} 
        & {\color{higher}+2.3} & {\color{lower}-0.9} & {\color{lower}-0.1} \\
        
        \sbest{w. \vocada-CLIP}
        & \sbest{30.4} & \sbest{52.4} & \sbest{46.7}
        & \sbest{35.3} & \sbest{52.2} & \sbest{47.8} 
        & \sbest{33.9} & \sbest{54.4} & \sbest{49.0} 
        & \sbest{\color{higher}+3.1} & \sbest{\color{higher}+1.6} & \sbest{\color{higher}{+2.0}} \\

        \fbest{w. \vocada-LLM}
        & \fbest{30.6} & \fbest{52.9} & \fbest{47.1} 
        & \fbest{35.5} & \fbest{52.6} & \fbest{48.1} 
        & \fbest{34.1} & \fbest{54.7} & \fbest{49.3} 
        & \fbest{\color{higher}+3.3} & \fbest{\color{higher}+2.0} & \fbest{\color{higher}+2.4} \\
        
        \cmidrule{1-1}
        \cmidrule(lr){2-4}
        \cmidrule(lr){5-7}
        \cmidrule(lr){8-10}
        \cmidrule(lr){11-13}

        \color{upperboundcolor}w. 
        \vocada-Oracle
        & \color{upperboundcolor}{33.8} & \color{upperboundcolor}{57.0} & \color{upperboundcolor}{50.9} 
        & \color{upperboundcolor}{39.3} & \color{upperboundcolor}{56.9} & \color{upperboundcolor}{52.3} 
        & \color{upperboundcolor}{37.8} & \color{upperboundcolor}{59.5} & \color{upperboundcolor}{53.8} 
        & \color{upperboundcolor}{+6.9} & \color{upperboundcolor}{+6.4} & \color{upperboundcolor}{+6.5} \\
        \bottomrule
    \end{tabular}

    \lbltab{main_coco}
    \lesspace
\end{table*}
\begin{table}[h]
    \centering
    \small
    \setlength{\tabcolsep}{4.8pt} 
    \caption{
    \textbf{Comparison on CDTE-Objects365}. We apply \vocada to Detic and CoDet with a Swin-B~\cite{liu2021swin} backbone.
    }
    \vspace{-2.0mm}
    \begin{tabular}{lcccccc}
        \toprule
        \multicolumn{1}{c}{\multirow{2}{*}{Method}}
        & \multicolumn{2}{c}{Detic} 
        & \multicolumn{2}{c}{CoDet}
        & \multicolumn{2}{c}{$\overline{\Delta}$} \\

        \cmidrule(lr){2-3}
        \cmidrule(lr){4-5}
        \cmidrule(lr){6-7}

        & $\text{AP}$ & $\text{AP}_{50}$ 
        & $\text{AP}$ & $\text{AP}_{50}$ 
        & $\text{AP}$ & $\text{AP}_{50}$ \\


        \cmidrule{1-1}
        \cmidrule(lr){2-3}
        \cmidrule(lr){4-5}
        \cmidrule(lr){6-7}

        \rowcolor{vanillacolor}{Baseline}
        & {21.5} & {29.5} 
        & {14.3} & {21.9}  
        & {-} & {-} \\

        \cmidrule{1-1}
        \cmidrule(lr){2-3}
        \cmidrule(lr){4-5}
        \cmidrule(lr){6-7}

        w. RAM++
        & {17.9} & {23.3} 
        & {11.4} & {17.1} 
        & {\color{lower}-3.2} & {\color{lower}-5.5}\\

        \cmidrule{1-1}
        \cmidrule(lr){2-3}
        \cmidrule(lr){4-5}
        \cmidrule(lr){6-7}

        w. \vocada-SBert
        & {21.7} & {29.9}
        & {23.1} & {22.2}
        & {\color{higher}+0.4} & {\color{higher}+0.3}\\

        \sbest{w. \vocada-CLIP}
        & \sbest{22.5} & \sbest{30.3} 
        & \sbest{16.0} & \sbest{22.3}  
        & \sbest{\color{higher}+1.3} & \sbest{\color{higher}+0.6}  \\
        
        \fbest{w. \vocada-LLM}
        & \fbest{23.4} & \fbest{32.1} 
        & \fbest{16.8} & \fbest{23.4} 
        & \fbest{\color{higher}+2.2} & \fbest{\color{higher}+2.1} \\
        
        \cmidrule{1-1}
        \cmidrule(lr){2-3}
        \cmidrule(lr){4-5}
        \cmidrule(lr){6-7}

        \color{upperboundcolor}w. \vocada-Oracle
        & \color{upperboundcolor}{35.2} & \color{upperboundcolor}{50.3}  
        & \color{upperboundcolor}{25.9} & \color{upperboundcolor}{39.7} 
        & \color{upperboundcolor}{+12.7} & \color{upperboundcolor}{+19.3} \\
        \bottomrule
    \end{tabular}
    \lesspace
    \lbltab{main_objects365}
\end{table}

\section{Experiments}
\lblsec{ovod_expt}

\myparagraph{Benchmarks.}
We evaluate \vocada on COCO~\cite{lin2014microsoft} and Objects365~\cite{shao2019objects365}. For COCO, we follow the open-vocabulary evaluation (OVE) protocol~\cite{zhu2023survey}, splitting the 80 classes into 48 base (seen) and 17 novel (unseen) while excluding 15 classes lacking WordNet~\cite{fellbaum1998wordnet} synsets. We train \ovod detectors on box-labeled base classes and evaluate them on 5$k$ test images containing both base and novel classes. For Objects365, we use the cross-dataset transfer evaluation (CDTE) protocol~\cite{zhou2022detecting, zhu2023survey}, training on box-labeled LVIS~\cite{gupta2019lvis} and then testing in a zero-shot manner on 80$k$ images of novel classes in Objects365. We report mAP averaged over multiple IoU thresholds and $\text{AP}_{50}$ at an IoU threshold of 0.5. Specifically, $\text{AP}^{novel}$, $\text{AP}^{base}$, and $\text{AP}^{all}$ are computed for novel, base, and all classes, respectively. Further details are in \refapp{supp_metrics}.

\myparagraph{{Comparisons}.}
We apply the proposed \vocada to three \ovod detectors: Detic~\cite{zhou2022detecting}, VLDet~\cite{lin2022learning}, and CoDet~\cite{ma2024codet}. Their vanilla versions 
{define}
our \textit{Baseline}. We also compare \vocada with the open-set tagger \textit{RAM++}~\cite{huang2023open}, which can replace \vocada by tagging $\widetilde{\classes}_I$ from the user-defined vocabulary $\classes$. Lastly, we design an \textit{Oracle} version of \vocada by using the test image’s ground-truth 
classes as $\widetilde{\classes}_I$, which represents the optimal adapted vocabulary for each image.

\subsection{Main Results}
\lblsec{expt_main}

\myparagraph{Oracle validation.}
\reftab{main_coco} and~\reftab{main_objects365} show the large gaps between the Baseline (first row) and Oracle performance (last row): up to {\color{higher}\textbf{+6.9}} points on OVE-COCO and {\color{higher}\textbf{+19.3}} points on CDTE-Objects365. This confirms that detection performance can be 
{significantly}
improved when the vocabulary is well-adapted to the given image. This validates our core idea, namely that discarding non-relevant (distracting) classes from the vocabulary improves detection performance.

\myparagraph{Quantitative evaluation of \vocada.}
We evaluate \vocada along two dimensions: \textit{i)} 
{improvements}
to baseline performance, and \textit{ii)} 
generalization across detectors.
In \reftab{main_coco} and~\reftab{main_objects365}, {we show that} \vocada yields notable average gains of up to {\color{higher}\textbf{+3.3}} and {\color{higher}\textbf{+2.2}} points, respectively, consistent across different detectors. 
This confirms the effectiveness and generalizability of our \textit{plug-and-play} approach. 
\vocada also outperforms RAM++, highlighting the strength of our language-based pipeline.


\begin{figure}[t]
  \includegraphics[width=0.98\linewidth]{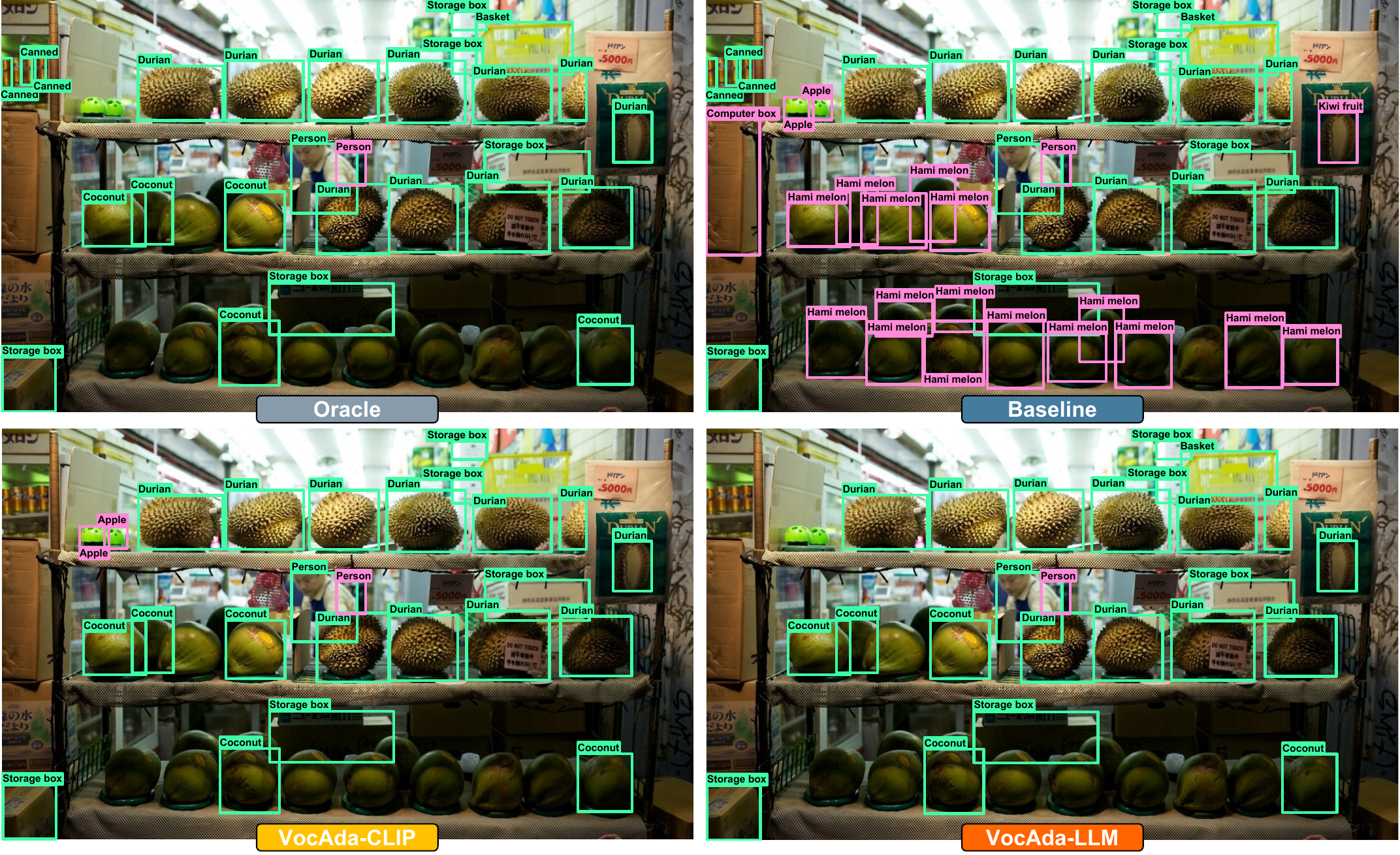}
  \vspace{-2.0mm}
    \caption{
    \textbf{Qualitative Comparison on Objects365.}
    We use Detic (Swin-B backbone) trained on LVIS and ImageNet-21k as the \ovod detector.
    Correct and incorrect detections appear in {\color{correct_dts}green} and {\color{wrong_dts}pink}, respectively, with a 0.5 confidence threshold.
    }
  \vspace{-6.0mm}
  \lblfig{main_qualitative}
\end{figure}
\myparagraph{Qualitative results.}
We present qualitative comparisons of Oracle, Baseline, \vocada-CLIP, and \vocada-LLM in \reffig{main_qualitative}. 
As observed, the baseline, which uses the entire vocabulary, easily confuses distracting classes (\eg, labeling a ``Storage Box'' as a ``Computer Box''). 
\vocada alleviates this by adapting the vocabulary to the image’s semantic context.

\myparagraph{\vocada{} variants.}
We study three variants of our method based on different Class Selectors: \vocada-LLM (LLM-based), \vocada-CLIP (CLIP-based), and \vocada-SBert (using Sentence-BERT~\cite{reimers2019sentence} instead of CLIP). As \reftab{main_coco} and \reftab{main_objects365} show, \vocada-LLM consistently outperforms the others, suggesting that the LLM better interprets semantic context and thereby more accurately identifies relevant categories. \vocada-CLIP also provides satisfactory gains over \vocada-SBert, likely due to CLIP’s stronger visual-semantic alignment.
While \vocada-LLM is more accurate, \vocada-CLIP avoids the computational costs of LLMs, as analyzed in ~\refapp{additional_expt}.

\subsection{Analysis}
\lblsec{expt_analysis}

\begin{table}[!t]
    \centering
    \small
    \setlength{\tabcolsep}{4.1pt} 
    \caption{
    \textbf{Study of different top-k selection} of \vocada-CLIP on OVE-COCO. We use Detic and compare different k-values used for selecting classes from the user-defined vocabulary.
    }
    \vspace{-3.0mm}
    \begin{tabular}{lcccccc}
        \toprule

        \multicolumn{1}{c}{Method}
        & $\text{AP}_{50}^{novel}$ & $\text{AP}_{50}^{base}$ & $\text{AP}_{50}^{all}$
        & $\text{AP}_{50}^{novel}$ & $\text{AP}_{50}^{base}$ & $\text{AP}_{50}^{all}$ \\


        \cmidrule{1-1}
        \cmidrule(lr){2-4}
        \cmidrule(lr){5-7}
        
        \rowcolor{vanillacolor}{Baseline}
        & {27.8} & {51.1} & {45.0} 
        & {-} & {-} & {-} \\

        \cmidrule{1-1}
        \cmidrule(lr){2-4}
        \cmidrule(lr){5-7}

        \fbest{top-k=1}
        & \fbest{30.4} & \fbest{52.4} & \fbest{46.7} 
        & \fbest{\color{higher}+2.6} & \fbest{\color{higher}+1.3} & \fbest{\color{higher}+1.7} \\
        
        top-k=2
        & {29.6} & {52.1} & {46.2}
        & {\color{higher}+1.8} & {\color{higher}+1.0} & \color{higher}{+1.2} \\

        top-k=3
        & {29.1} & {51.9} & {45.9}
        & {\color{higher}+1.4} & {\color{higher}+0.8} & \color{higher}{+0.9} \\

        \bottomrule
    \end{tabular}

    \lbltab{study_topk}
    \vspace{-2.00mm}
\end{table}

\begin{figure}[!t]
  \includegraphics[width=0.98\linewidth]{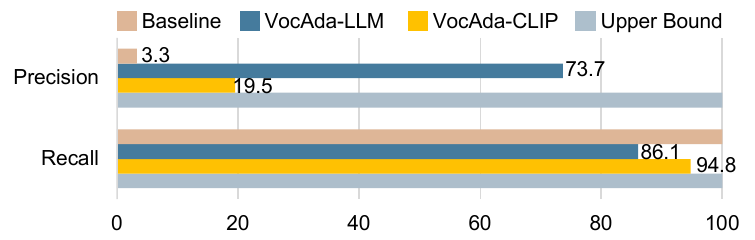}
  \vspace{-4.0mm}
  \caption{
  \textbf{
  Vocabulary adaptation quality} on OVE-COCO, measured via average precision and recall.
  }
  \lblfig{comp_category_quality}
  \vspace{-5.0mm}
\end{figure}
\myparagraph{Quality of the adapted vocabulary.}
Next, we evaluate our two best methods in terms of vocabulary adaptation quality in \reffig{comp_category_quality}. 
For an image $\bI$ and vocabulary $\classes$, we define classes actually appearing in $\bI$ as true positives (TPs).
A well-adapted vocabulary $\widetilde{\classes}_I$ should \textit{i)} miss few{er} relevant classes (minimize false negatives, FNs) and \textit{ii)} include few{er} irrelevant classes (minimize false positives, FPs). 
We quantify this using Precision = $TPs / (TPs + FPs)$ and Recall = $TPs / (TPs + FNs)$. 
\reffig{comp_category_quality} reports these measures on OVE-COCO. 
The Baseline (full vocabulary) naturally has perfect recall but no filtering, while the Oracle (ground-truth classes) achieves both perfect recall and precision. 
Among our two top methods, \vocada-LLM exhibits higher precision than \vocada-CLIP (73.7\% \textit{vs.}\ 19.5\%), indicating better removal of distractors. 
However, \vocada-CLIP attains higher recall (94.8\% vs.\ 86.1\%), indicating it retains more relevant classes.


\myparagraph{Varying the number of selected classes.}
\vocada-CLIP selects the most similar class name per noun by default, i.e., $k=1$. Increasing $k$ expands $\widetilde{\classes}_I$ and reduces false negatives but risks more false positives. 
\reftab{study_topk} shows that larger $k$ lowers {the gain} over the Baseline, indicating that the benefits of fewer missed classes are outweighed by {adding} more distractors.

\myparagraph{Sensitivity to different \llm{s}.} When we replace Llama3-8B in \vocada-LLM with either Mistral-7B-Instruct~\cite{jiang2023mistral} or GPT-3.5~\cite{chatgpt}, \vocada-LLM remains effective regardless of the \llm{} used, as shown in \reftab{study_llm}. This is likely because its class selection task is straightforward for modern LLMs.

\begin{table}[!t]
    \centering
    \small
    \setlength{\tabcolsep}{3.5pt} 
    \caption{
    \textbf{Different \llm{s} in \vocada-LLM}, using the Detic detector and the OVE-COCO benchmark.
    }
    \vspace{-3.0mm}
    \begin{tabular}{lcccccc}
        \toprule

        \multicolumn{1}{c}{Method}
        & $\text{AP}_{50}^{novel}$ & $\text{AP}_{50}^{base}$ & $\text{AP}_{50}^{all}$
        & $\text{AP}_{50}^{novel}$ & $\text{AP}_{50}^{base}$ & $\text{AP}_{50}^{all}$ \\


        \cmidrule{1-1}
        \cmidrule(lr){2-4}
        \cmidrule(lr){5-7}
        
        \rowcolor{vanillacolor}{Baseline}
        & {27.8} & {51.1} & {45.0} 
        & {-} & {-} & {-} \\

        \cmidrule{1-1}
        \cmidrule(lr){2-4}
        \cmidrule(lr){5-7}
        
        Mistral-7B
        & {30.1} & {52.3} & {46.5}
        & {\color{higher}+2.3} & {\color{higher}+1.2} & \color{higher}{+1.5} \\

        GPT-3.5
        & 30.5 & 53.1 & 47.2
        & \color{higher}+2.7 & \color{higher}+2.0 & \color{higher}+2.2 \\

        Llama3-8B
        & 30.6 & 52.9 & 47.1
        & \color{higher}+2.8 & \color{higher}+1.8& \color{higher}+2.1 \\

        \bottomrule
    \end{tabular}

    \lbltab{study_llm}
    \vspace{-2.0mm}
\end{table}

\begin{figure}[!t]
  \includegraphics[width=0.98\linewidth]{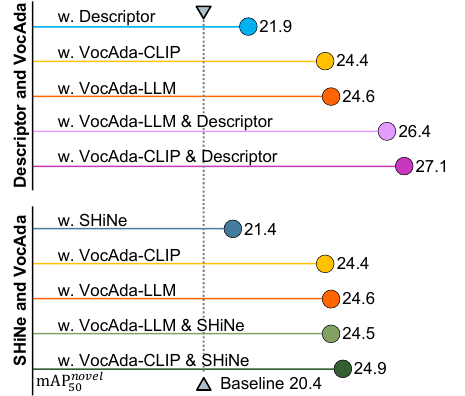}
  \vspace{-2.0mm}
    \caption{
    \textbf{Comparison with prompt engineering methods on OVE-COCO.}
    We use Detic as our Baseline and compare \vocada with two prompt-engineering approaches, Descriptor and \shine, including their combinations with \vocada.
    }
  \lblfig{comp_prompt_eng}
  \vspace{-6.0mm}
\end{figure}
\myparagraph{\vocada with prompt engineering.}
We compare and combine \vocada with two prompt-engineering methods: Descriptor~\cite{menon2022visual} and \shine~\cite{liu2024shine}. 
Descriptor uses GPT-3 to gather class descriptions for each $c_j \in \classes$ and augments $\text{T}(\cdot)$ with sentences mentioning the target class and its descriptions. 
\shine augments $\text{T}(\cdot)$ with super-/sub-categories using a semantic hierarchy. 
Both methods refine the textual prompt $\text{T}(\cdot)$, so they can be integrated with \vocada.
As shown in \reffig{comp_prompt_eng}, \vocada alone yields larger gains than Descriptor or \shine individually. 
Combining \vocada and Descriptor further improves performance, while combining \vocada with \shine yields only minor additional gains. 

\section{Conclusion}
\lblsec{conclusion}
Open-vocabulary object detection (\ovod) 
{allows}
users {to} freely define the vocabulary for a detector
, making vocabulary quality pivotal to performance. Inspired by our Oracle results—where removing distracting classes yields large gains—we introduce \vocada, a \textit{plug-and-play} module for \textit{off-the-shelf} \ovod detectors. Our experiments show that \vocada leverages \vlm{} and \llm{} components to adapt the vocabulary at test time and consistently improves performance across multiple detectors and benchmarks. Further analyses reveal promising directions for future work on vocabulary adaptation, especially in increasing precision while maintaining recall.
\vspace{1.5 mm}

\noindent\small{\textbf{Acknowledgements.}~E.R. is supported by MUR PNRR project FAIR - Future AI Research (PE00000013), funded by NextGenerationEU and EU projects SPRING (No. 871245) and ELIAS (No. 01120237). M.L. is supported by the PRIN project LEGO-AI (Prot. 2020TA3K9N). M.L. thanks Margherita Potrich for her constant
support.
}

\clearpage

\vfill\pagebreak



\bibliographystyle{IEEEbib}
\bibliography{refs}

\clearpage
\maketitlesupplementary
\lblsec{appendix}
\appendix

In this supplementary material, we first presents further experimental studies, including the impact of synonyms on VocAda-LLM and VocAda-CLIP, as well as a computational cost study of the proposed method.
Next, we detail the evaluation metrics in \refapp{supp_metrics}. Lastly, \refapp{supp_prompt} provides the complete prompts for the Image Captioner (\textbf{IC}) and the LLM-based Class Selector (CS), along with their design details. We will publicly release our code and the intermediate results upon publication.

\section{Further Studies}
\lblsec{additional_expt}
\begin{table}[!ht]
    \centering
    \small
    \setlength{\tabcolsep}{2.0pt} 
    \caption{
    \textbf{Influence of synonyms on \vocada-LLM}, using the Detic detector and the OVE-COCO benchmark.
    }
    \begin{tabular}{lcccccc}
        \toprule

        \multicolumn{1}{c}{Method}
        & $\text{AP}_{50}^{novel}$ & $\text{AP}_{50}^{base}$ & $\text{AP}_{50}^{all}$
        & $\text{AP}_{50}^{novel}$ & $\text{AP}_{50}^{base}$ & $\text{AP}_{50}^{all}$ \\


        \cmidrule{1-1}
        \cmidrule(lr){2-4}
        \cmidrule(lr){5-7}
        
        \rowcolor{vanillacolor}{Baseline}
        & {27.8} & {51.1} & {45.0} 
        & {-} & {-} & {-} \\

        \cmidrule{1-1}
        \cmidrule(lr){2-4}
        \cmidrule(lr){5-7}

        w/o Synonyms
        & {30.5} & {49.5} & {44.5}
        & {\color{higher}+2.7} & {\color{lower}-1.6} & \color{lower}{-0.5} \\
        
        \fbest{w/ Synonyms}
        & \fbest{30.6} & \fbest{52.9} & \fbest{47.1} 
        & \fbest{\color{higher}+2.8} & \fbest{\color{higher}+1.8} & \fbest{\color{higher}+2.1} \\

        \bottomrule
    \end{tabular}
    \lbltab{study_synonyms}
\end{table}
\myparagraph{Influence of synonyms on \vocada-LLM.}
Including synonyms for class names in the user-defined vocabulary when prompting the
LLM-based \textbf{CS} module is a simple but crucial design choice, as shown in \reftab{study_synonyms}. During our initial exploration, we found, for example, that without synonyms some large and obvious objects like ``Couch" or ``TV'' were often missed by the LLM-based \textbf{CS} of \vocada, even though they were included in the image descriptions. This occurred because these categories were phrased differently in the captions (\eg, ``Sofa'' or ``Television'') and, hence, the LLM-based \textbf{CS} processed them as not relevant and discarded them. Including synonyms as cues in the system prompt of the LLM prevents this erroneous filtering, resulting in superior performance.

\myparagraph{Influence of synonyms on \vocada-CLIP.}
Using synonyms in the CLIP-based \textbf{CS} is less straightforward. Implementing synonyms for nouns would require querying an LLM at test time for each noun phrase, adding significant extra cost. We experimented with using synonyms for the class names queried offline, but this had no effect on the \vocada-CLIP results.


\begin{table}[!ht]
    \centering
    \small
    \caption{
    \textbf{Study of Computational Cost.}
    We evaluate the inference time and computational cost of the baseline detector (Detic Swin-B) and its integrations with VocAda-CLIP and VocAda-LLM on a Tesla V100 (32GB) using the COCO-val dataset with a batch size of 1. VocAda-LLM employs LLaVA-Next-7B~\cite{liu2024llavanext} as the VLM and Llama3-8B~\cite{meta2024llama3} as the LLM, while VocAda-CLIP uses CLIP ViT-L/14.
    }

    \begin{tabular}{lrr}
         \toprule
         Methods & Speed (sec/img) & GPU Requirement \\
         \cmidrule{1-1}
         \cmidrule{2-2}
         \cmidrule{3-3}
         Detic & 0.115 & 8 GB \\
         \cmidrule{1-1}
         \cmidrule{2-2}
         \cmidrule{3-3}
         Detic w. VocAda-CLIP & 5.572 & 17 GB\\
         Detic w. VocAda-LLM & 10.699 & 31 GB\\
         \bottomrule
    \end{tabular}
    \lesspace
    \lbltab{computational_cost}
\end{table}

\myparagraph{Study of computational cost.}
\reftab{computational_cost} compares inference speed and computational requirements. Metrics were measured on a Tesla V100 (32GB) using Detic Swin-B~\cite{zhou2022detecting} as the detector, LLaVA-Next-7B~\cite{liu2024llavanext} as the VLM, Llama3-8B~\cite{meta2024llama3} as the LLM, and CLIP ViT-L/14~\cite{radford2021learning}, with the COCO-val~\cite{lin2014microsoft} dataset. Although inference time increases, our method requires manageable resources (17GB for VocAda-CLIP and 31GB for VocAda-LLM), making it suitable for real-world applications. While VLMs and LLMs do slow down the vanilla detection pipeline, this can be mitigated with advanced deployment strategies like TensorRT or SGLang, which can speed up LLaVA and Llama3 by 6X and 2X, respectively.

Importantly, we believe that current computational limits should not hinder exploring new paradigms. Large VLMs and LLMs are increasingly integrated into detection pipelines and co-run with detectors in applications like autonomous driving. In such systems, VocAda adds minimal overhead by utilizing existing VLM outputs (captions).

\myparagraph{Additional qualitative results.} We present additional qualitative comparison of Oracle, Baseline, VocAda-CLIP and VocAda-LLM in \reffig{additional_qualitative}. As observed, the baseline detector using the full vocabulary is
easily confused by distracting classes, incorrectly
classifying a “Curling” on a sports court
as a “Teapot”. VocAda alleviates this confusion by
adapting the vocabulary to the input image based
on its interpretation of the semantic context. Even
when VocAda does not lead to a correct detection,
at least it avoids a mis-detection (see the curling
stone in the right panel).

\begin{figure}[!t]
  \includegraphics[width=0.98\linewidth]{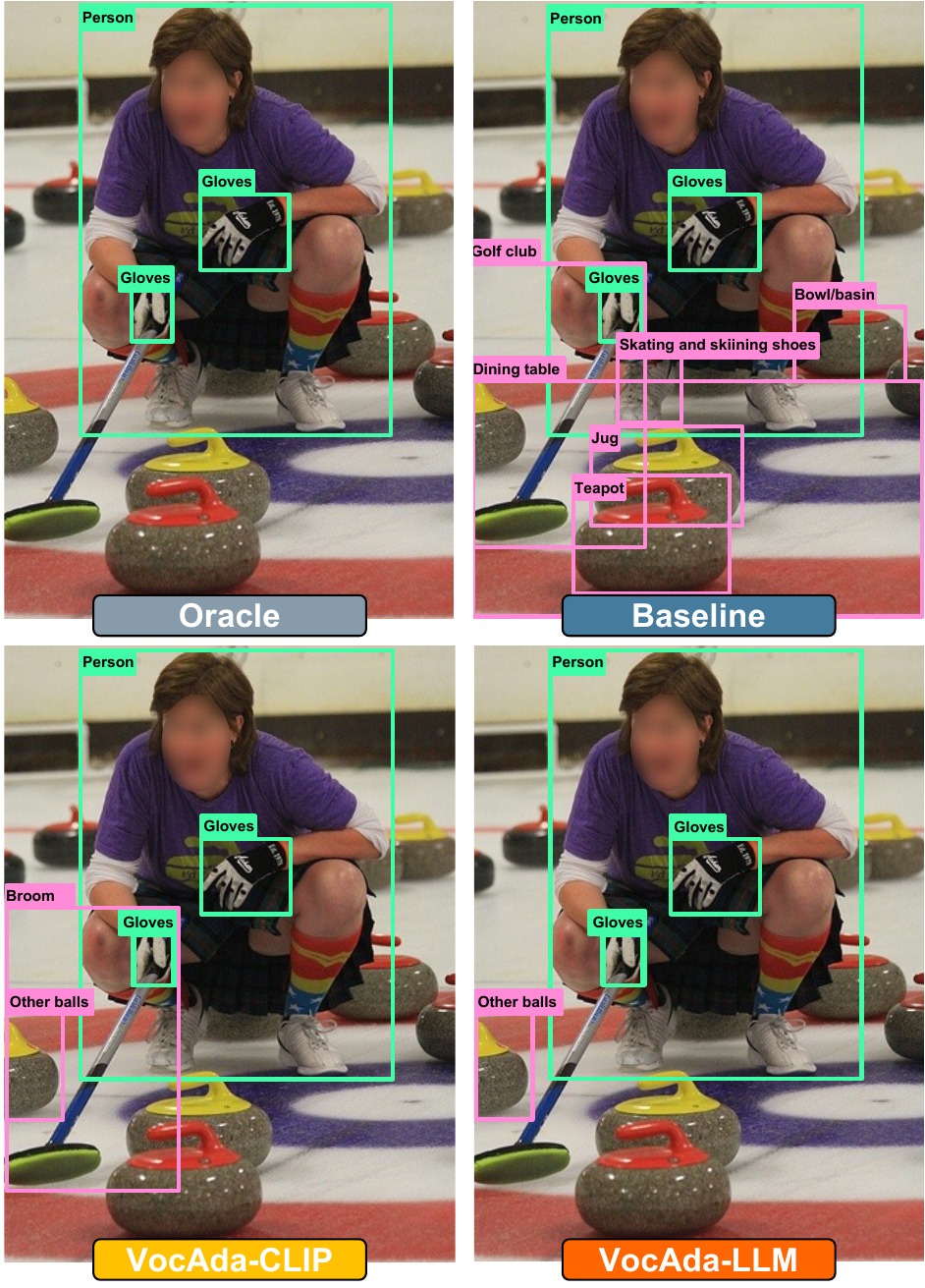}
  \caption{
    \textbf{Additional Qualitative Results on Objects365.}
    We use Detic (Swin-B backbone) trained on LVIS and ImageNet-21k as the \ovod detector.
    Correct and incorrect detections appear in {\color{correct_dts}green} and {\color{wrong_dts}pink}, respectively, with a 0.5 confidence threshold.
  }
  \lblfig{additional_qualitative}
\end{figure}


\section{Details on the Evaluation Metrics}
\lblsec{supp_metrics}
\begin{figure}[!h]
  \includegraphics[width=0.98\linewidth]{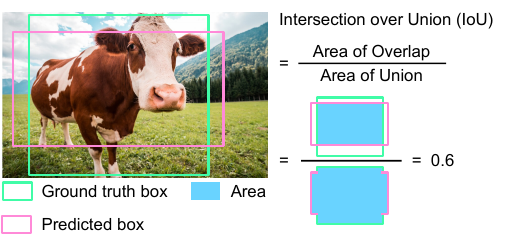}
  \caption{
  {Example of Intersection over Union (IoU) calculation.}
  }
  \lblfig{example_iou}
\end{figure}
As illustrated in \reffig{example_iou}, given a predicted bounding box and the closest ground truth box, the Intersection over Union (IoU) is the ratio of their intersection area to their union area. For each object class, predictions are sorted by their confidence scores in descending order, and Average Precision (AP) is calculated as the area under the precision-recall curve. This combines precision and recall to provide a single performance measure for detection tasks. 

Mean Average Precision (mAP) is the mean of the AP values, averaged across novel (unseen), base (seen), or all classes, denoted by $\text{AP}^{novel}$, $\text{AP}^{base}$, and $\text{AP}^{all}$, respectively. $\text{AP}_{50}$ refers to mAP when IoU is considered with a threshold of 0.5. Otherwise, AP values are computed for thresholds from 0.5 to 0.95 in steps of 0.05 and then averaged.

Taking the calculation of $\text{AP}_{50}$ as an example, we start by computing the IoU for each predicted bounding box and ground truth pair. A prediction is considered a True Positive (TP) if: \textit{i)} its IoU is 0.50 or higher, and \textit{ii)} its predicted class label matches the ground truth; otherwise, it's a False Positive (FP). Detections are sorted by confidence scores in descending order, and for each prediction, we evaluate its IoU and class label against the ground truth. Precision and recall are calculated at each detection: precision is the ratio of TPs to the total number of predictions (TPs + FPs), and recall is the ratio of TPs to the total number of ground truth objects (TPs + FNs). These values are used to plot the precision-recall curve, and the area under this curve represents the $\text{AP}_{50}$ measurement. The final $\text{AP}_{50}$ is averaged across all evaluated classes, summarizing the model's performance in terms of both localization and classification for the test dataset.

\section{Prompt Details}
\lblsec{supp_prompt}
In this section, we provide the complete prompts used in \vocada and elaborate on the design choices for prompting.

\begin{figure*}[!t]
  \begin{center}
  \includegraphics[width=0.95\linewidth]{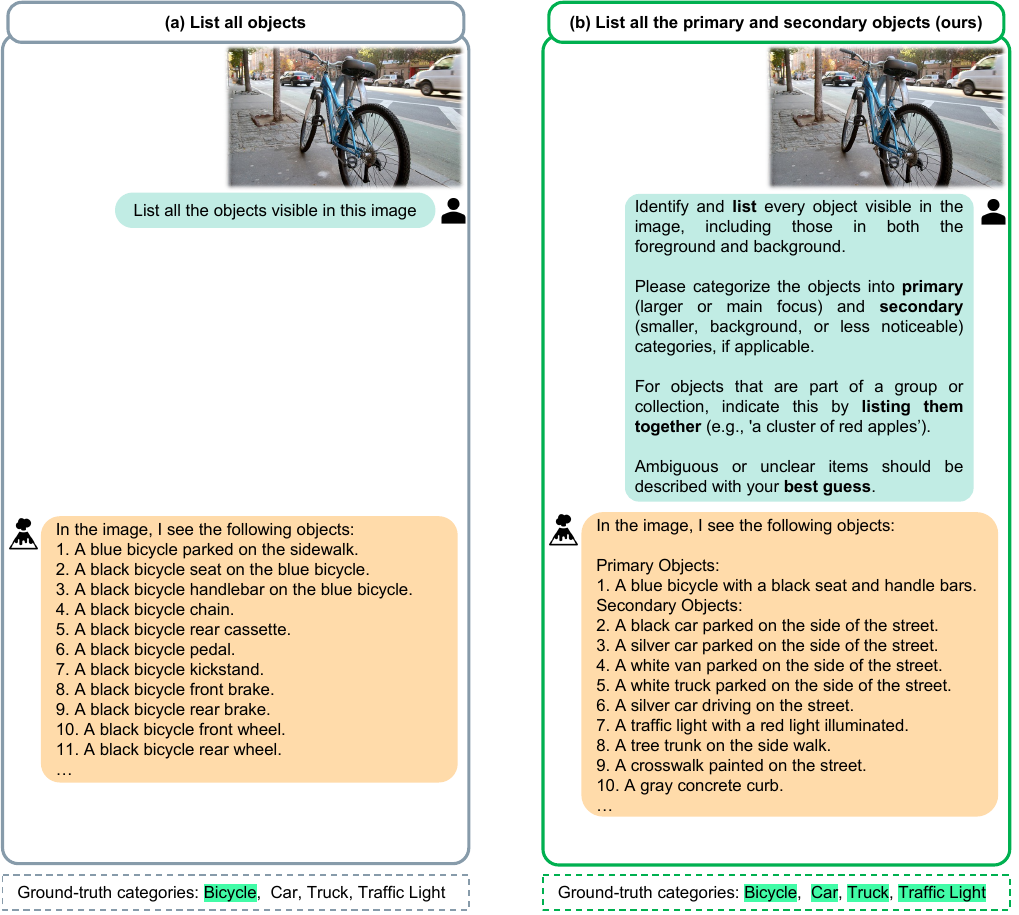}
   \end{center}
  \caption{
  Comparison of image captioner prompts. (a) A simple prompt is used to list all the visible objects in the image. (b) We design a better prompt to improve the comprehensiveness of the description by instructing the captioner to list primary and secondary objects. The ground-truth categories that are mentioned by the output caption are highlighted in {\colorbox{promptGreen}{Green}}.
  }
  \lblfig{fig_prompt_ic}
\end{figure*}

\subsection{Prompting Image Captioner (IC)}
\lblsec{prompt_ic}

The comprehensiveness of the description generated by the \textbf{IC} is crucial for the subsequent steps of \vocada. The image description should capture as many categories present in the current image as possible. Even state-of-the-art VLMs often neglect background objects in images, focusing on more prominent foreground objects when prompted with a simple prompts such as ``List all the objects visible in this image''. For instance, as shown in \reffig{fig_prompt_ic}(a), although the cars and trucks in the background are clearly visible, the \textbf{IC} only describes the foreground object ``bicycle''. To address this, as shown in \reffig{fig_prompt_ic}(b), we propose a prompt strategy that instructs the \textbf{IC} to not only list all visible objects but also categorize them into \textit{primary} and \textit{secondary} groups. Even though we do not need the grouping results \textit{per se}, this technique effectively guides the \textbf{IC} to comprehensively describe both large and focused foreground objects (primary) and small and background objects (secondary), such as ``Traffic Light'' in \reffig{fig_prompt_ic}(b).

In \reffig{fig_prompt_ic}(b), we show the full prompt used for the \textbf{IC} (LLaVA-Next-7B~\cite{liu2024llavanext}) in \vocada to describe the image, creating textual measurements of the objects visible in the image.

In addition, there are two design choices worth mentioning. First, in our prompt, we instruct the \textbf{IC} to list a group of objects together (\eg, ``a cluster of red apples'') instead of one by one. This technique prevents the \textbf{IC} from generating \textit{repetitive} patterns, which are lengthy and not useful for the following steps. The goal of the \textbf{IC} is to comprehensively capture object categories likely to appear in the current images. Therefore, we further ask the \textbf{IC} to provide ``best guesses'' for unclear items in the prompt. This design force the IC to reason possible objects that might be present in the image based on its interpretation. While this might introduce extra noise, the Class Selector module can alleviate most of them, especially if they are unrelated to the global image context. In \vocada, we use the exact prompt shown in \reffig{fig_prompt_ic}(b) for all experiments.

\begin{figure*}[!ht]
  \begin{center}
  \includegraphics[width=\linewidth]{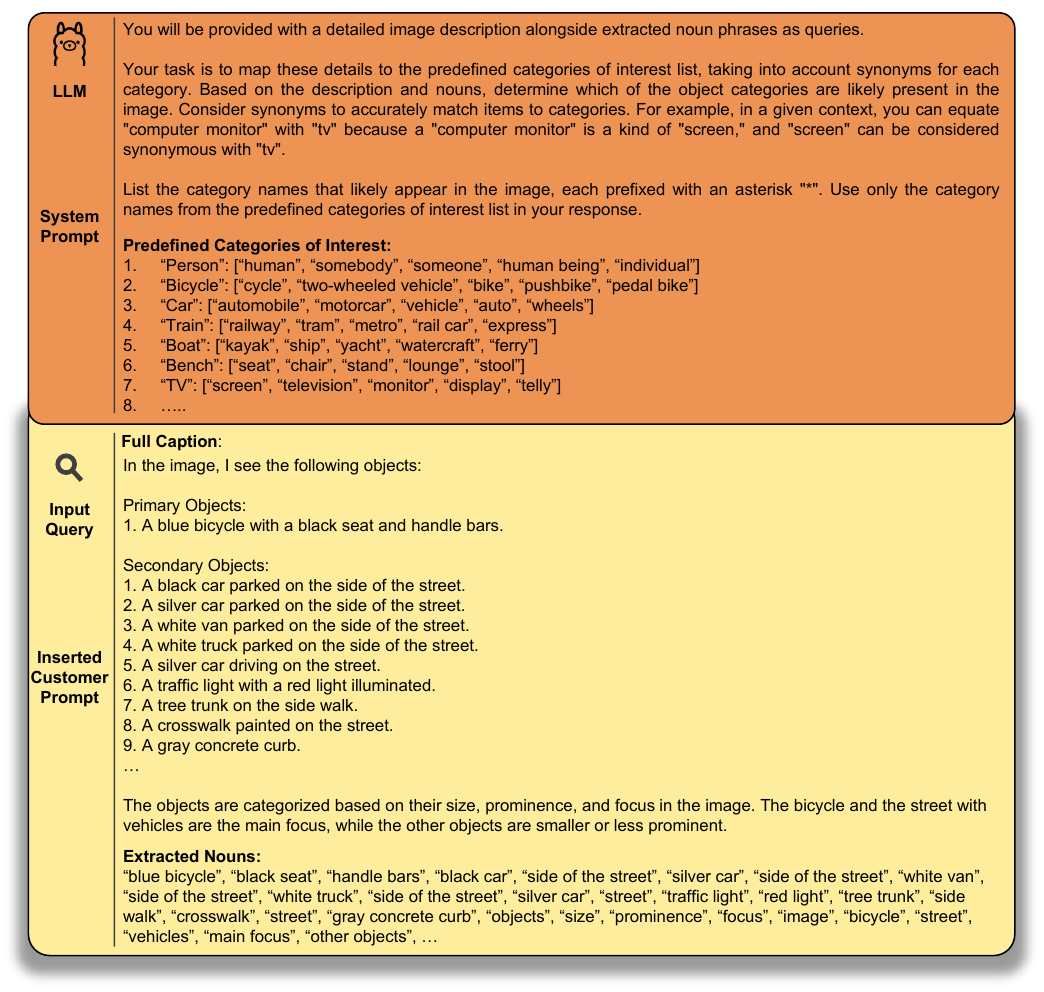}
  \end{center}
  \caption{
  \textbf{Complete prompts used the \llm{} Proposal based Class Selector (CS).} \textbf{Top}: The system prompt includes the user defined categories enriched with a set of synonyms and the task instruction. The latter guides the \llm{} to select from the category list the ones that are relevant given as input an image description and the set of extracted noun phrases. This system prompt is used to instantiate the \llm{} agent as the \textbf{CS}. \textbf{Bottom}: During inference, the full image description, ($\captions_{I}$ provided by the \textbf{IC}) alongside the extracted noun phrases ($\nouns_{I}$ from the \textbf{NE})  are fed to the system as customer prompt input. Subsequently, the \llm{} automatically propose the selected category names based on this input.  
  }
  \lblfig{fig_prompt_llm}
\end{figure*}

\subsection{Prompting \llm{} as Class Selector (CS)}
\lblsec{prompt_llm}

In \reffig{fig_prompt_llm}, we present the complete system and customer prompts used for the \llm{} Proposal-based \textbf{CS} in \vocada. Specifically, we first instantiate a \llm{} agent, such as Llama3-8B~\cite{meta2024llama3}, with a system prompt that includes a task instruction and the user-defined vocabulary with their synonyms. The task instruction specifies the input query, the generated image caption $\captions_{I}$ provided by the \textbf{IC} and the corresponding extracted noun phrases $\nouns_{I}$ from the \textbf{NE}, that the \llm{} will receive during inference. It then guides the \llm{} with a detailed task description, which is to select relevant categories likely to appear in the image from the embedded user-defined vocabulary based on the input, taking also synonyms into consideration. Subsequently, the \llm{} is instructed with the output format of the selected categories (prefixing each category name with an asterisk ``*'') for easier post parsing. This \llm{} instantiation is conducted before large scale inference.

Therefore, during inference, the \llm{}-based \textbf{CS} takes the complete image description $\captions_{I}$ and the corresponding noun phrases $\nouns_{I}$ as the user input without any additional instructions and automatically outputs a selected category set as $\widetilde{\classes}_{I}$.

\end{document}